\documentclass[letterpaper, 10 pt, conference]{ieeeconf}  

\IEEEoverridecommandlockouts                              

\usepackage{graphics} 
\usepackage{epsfig} 
\usepackage[T1]{fontenc}
\usepackage{xcolor,soul}
\usepackage{bm}
\usepackage{subfigure}
\usepackage{amsfonts,amssymb, amsmath}
\usepackage{cases}
\usepackage{graphicx}
\usepackage{booktabs}
\usepackage{multirow}
\usepackage{multicol}
\usepackage{flushend}
\usepackage{cite}
\usepackage[colorlinks,
            linkcolor=blue,
            anchorcolor=blue,
            citecolor=blue]{hyperref}
\hypersetup{colorlinks,
	allcolors=black,
	pdfstartview=Fit,
	breaklinks=true}
\newcounter{RNum}
\renewcommand{\theRNum}{\arabic{RNum}}
\newcommand{\Remark}{\noindent\textit{\textbf{Remark}~\refstepcounter{RNum}\textbf{\theRNum}: }}
\soulregister\Remark7
\title{\LARGE \bf
Integrated Behavior Planning and Motion Control for Autonomous Vehicles with Traffic Rules Compliance
}

\author{Haichao Liu, Kai Chen, Yulin Li, Zhenmin Huang, Jianghua Duan, and Jun Ma
   \thanks{This work was supported in part by the Department of Education of Guangdong Province under Grant 2023KQNCX101, and in part by the Project of Hetao Shenzhen-Hong Kong Science and Technology Innovation Cooperation Zone under Grant HZQB-KCZYB-2020083. Haichao Liu and Kai Chen contributed equally to this work. \textit{(Corresponding Author: Jun Ma.)}}
\thanks{Haichao Liu and Kai Chen are with the Robotics and Autonomous Systems Thrust, The Hong Kong University of Science and Technology (Guangzhou), Guangzhou, China (e-mail: hliu369@connect.ust.hk; kchen916@connect.hkust-gz.edu.cn).}
\thanks{Yulin Li and Zhenmin Huang are with the Division of Emerging Interdisciplinary Areas, The Hong Kong University of Science and Technology, Hong Kong SAR, China, and also with the Robotics and Autonomous Systems Thrust, The Hong Kong University of Science and Technology (Guangzhou), Guangzhou, China (e-mail: yline@connect.ust.hk; zhuangdf@connect.ust.hk).}
\thanks{Jianghua Duan is with the Department of Mechanical and Aerospace Engineering, The Hong Kong University of Science and Technology, Hong Kong SAR, China (e-mail: jhduan@ust.hk).}
 \thanks{Jun Ma is with the Robotics and Autonomous Systems Thrust, The Hong Kong University of Science and Technology (Guangzhou), Guangzhou, China, also with the Division of Emerging Interdisciplinary Areas, The Hong Kong University of Science and Technology, Hong Kong SAR, China, and also with the HKUST Shenzhen-Hong Kong Collaborative Innovation Research Institute, Futian, Shenzhen, China (e-mail: jun.ma@ust.hk).}
}

\begin{document}

\maketitle
\thispagestyle{empty}
\pagestyle{empty}

\begin{abstract}
In this article, we propose an optimization-based integrated behavior planning and motion control scheme, which is an interpretable and adaptable urban autonomous driving solution that complies with complex traffic rules while ensuring driving safety. Inherently, to ensure compliance with traffic rules, an innovative design of potential functions (PFs) is presented to characterize various traffic rules related to traffic lights, traversable and non-traversable traffic line markings, etc. These PFs are further incorporated as part of the model predictive control (MPC) formulation. In this sense, high-level behavior planning is attained implicitly along with motion control as an integrated architecture, facilitating flexible maneuvers with safety guarantees. Due to the well-designed objective function of the MPC scheme, our integrated behavior planning and motion control scheme is competent for various urban driving scenarios and able to generate versatile behaviors, such as overtaking with adaptive cruise control, turning in the intersection, and merging in and out of the roundabout. As demonstrated from a series of simulations with challenging scenarios in CARLA, it is noteworthy that the proposed framework admits real-time performance and high generalizability.
\end{abstract}

\section{Introduction}

\subsection{Background and Related Works}
Recent developments in the automobile industry prompt the emergence of autonomous vehicles, which enhance the safety and efficiency of transportation.
Typically, an autonomous driving system contains environment perception, behavior planning, motion planning and control technologies~\cite{qi2022hierarchical}. Despite the advancements in perceptions~\cite{prakash2021multi,deng2019cooperative}, a series of problems remain primarily unsolved in developing effective behavior planning and motion control schemes, especially in various complicated urban driving scenarios. 
With the rapid development of data science and artificial intelligence, learning-based methods are investigated to make decisions and generate control commands for autonomous vehicles. In~\cite{zhang2021enabling,chen2019model}, reinforcement learning is adopted to address difficult urban traffic scenarios based on the data recorded from the real world or reactive simulators. However, the lack of interpretability hinders their wide application in real-world applications. Another popular category is optimization-based methods, which also remove heuristic rules but provide good interpretability. For motion planning, iLQR is widely used to facilitate complex driving tasks and alleviate the high computational burden by applying advanced optimization techniques~\cite{ma2022alternating,ma2022local,Huang2023decentralized}. To advance behavior planning, integrated optimization frameworks typically unify the processes of behavior planning and low-level control by formulating them as an integrated optimization problem~\cite{guan2022integrated}. In ~\cite{sahin2020autonomous,eiras2021two}, mixed-integer programming (MIP) is adopted to handle the behavior planning task and obtain the discretized decision variables. However, MIP is non-convex and generally considered to be time-consuming. 
\begin{figure}[t]
    \centering
    \includegraphics[width=1\linewidth]{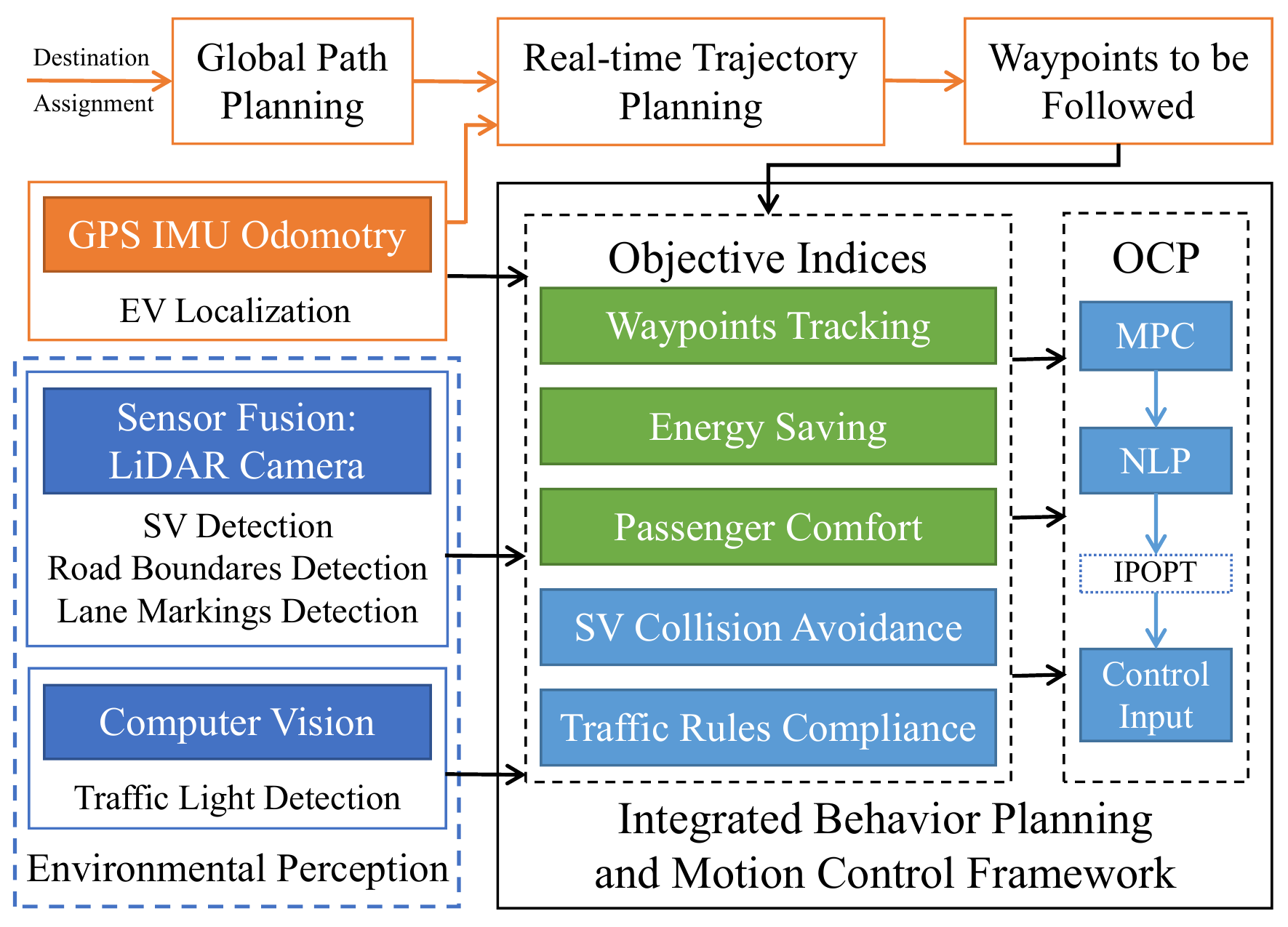}
    \caption{The proposed integrated behavior planning and motion control framework for autonomous driving. The framework models the objectives of collision avoidance and traffic rule compliance as potential functions.}
    \label{fig:idmc_framework}
\end{figure}

For the improvement of computational efficiency and driving safety, the artificial potential field (APF) can be suitably adopted, which integrates various forms of potential functions (PFs) into the optimization goal to realize simultaneous behavior planning and motion control, exhibiting both flexibility in problem formulation and high computation efficiency. Remarkably, a model predictive control scheme incorporating APF is proposed in~\cite{rasekhipour2016potential}, where vehicles are modeled as uncrossable obstacles and collision avoidance with them is successfully performed in a simple straight-line scenario. In~\cite{wang2019crash}, APF is introduced into emergency avoidance tasks, and the minimum collision cost is obtained when the collision cannot be avoided. In~\cite{ma2019active}, the lane-change maneuver is realized by applying an attractive force generated by the goal point. The above literature demonstrates the 
potential of APF for behavior planning in autonomous driving. Hence, \cite{zhang2022integrated} proposes an integrated framework with distributed APF for autonomous vehicles, and achieved a high success rate in a single-lane intersection. But this work only combines the traffic participants' APF and Depth-First-Search algorithm in the Frenet frame, which leads to low computational efficiency, especially in higher dimensional conditions, and this work is hard to generalize to other urban driving conditions with multiple road lanes. 

Furthermore, the above literature only assumes all the road lanes are the same and there is no non-traversable road lane marking. However, when the vehicle is in a tunnel, behind an intersection, or on a bridge, etc., lane changing is forbidden~\cite{yan2022cooperative}. Another common problem of the current literature is that traffic lights are not typically considered. For example, in~\cite{bae2019design}, the red traffic light state is treated as a hard constraint, which causes the vehicle to brake suddenly when it is close to the intersection; and apparently, this affects the comfort of passengers and even driving safety in dense traffic conditions. Therefore, it remains an interesting problem to propose specific strategies to cater to more traffic rules like different kinds of road lane markings and traffic control signals based on an effective design of the potential functions (PFs) of them. Moreover, finding an appropriate way to integrate APF into autonomous driving systems is valuable to improving computational efficiency.

\subsection{Contributions and Paper Structure}

This paper proposes an interpretable and computationally efficient optimization-based autonomous driving framework that realizes the integration of behavior planning and motion control with a global navigation path. The contributions are as follows:

\begin{itemize}
    \item  An integrated behavior planning and motion control scheme is proposed, which renders real-time performance in urban autonomous driving. 
\item  The proposed framework admits high generalizability due to 
an appropriate MPC problem formulation in the Cartesian frame, which is adaptable to 
varieties of urban driving conditions. 

\item An innovative design of PFs is proposed to characterize the traffic rules that are commonly encountered in urban driving scenarios, such that compliance with traffic rules is ensured.
\item Thorough simulations are performed in CARLA towards a series of urban driving scenarios, which verifies the effectiveness and efficiency of our proposed integrated behavior planning and motion control scheme.
\end{itemize}

The remainder of this paper is organized as follows: Section~II formulates the MPC problem with a non-linear vehicle dynamics model and the basic framework of our proposed scheme. In Section~III, the vehicles and traffic rules are modeled with APF to be incorporated into the MPC scheme. Section~IV shows the simulation results of our integrated behavior planning and motion control scheme for different scenarios. Finally, Section~V concludes this work.
\section{Problem Formulation and the Autonomous Driving Framework}

\subsection{Vehicle Dynamics Model}
Because the speed of autonomous vehicles varies widely under urban driving scenarios, this paper uses the nonlinear dynamics model proposed by~\cite{ge2021numerically} to represent the vehicle dynamics precisely. We utilize the discrete model derived from the backward Euler method:
\begin{align}
\label{bicycle_dynamics_equation}
\begin{split}
    & \bm x_{\tau+1}=f(\bm x_{\tau}, \bm u_{\tau})\\
    & =\left[\begin{array}{c}
    p_x(\tau)+T_s\left(v_x(\tau) \cos \varphi(\tau)-v_y(\tau) \sin \varphi(\tau)\right) \\
    p_y(\tau)+T_s\left(v_y(\tau) \cos \varphi(\tau)+v_x(\tau) \sin \varphi(\tau)\right) \\
    \varphi(\tau)+T_s \omega(\tau) \\
    v_x(\tau)+T_s a(\tau) \\
    \frac{m v_x(\tau) v_y(\tau)+T_s L_k \omega(\tau)-T_s k_f \delta(\tau) v_x(\tau)-T_s m v_x(\tau)^2 \omega(\tau)}{m v_x(\tau)-T_s\left(k_f+k_r\right)} \\
    \frac{I_z v_x(\tau) \omega(\tau)+T_s L_k v_y(\tau)-T_s l_f k_f \delta(\tau) v_x(\tau)}{I_z v_x(\tau)-T_s\left(l_f^2 k_f+l_r^2 k_r\right)}
    \end{array}\right]
\end{split},
\end{align}
where the state vector of the dynamics model is $\bm x = [\bm p, \varphi, \bm v, \omega]^T$ where $\bm p = [p_x, p_y]^T$ is the vehicle position in the global Cartesian frame, and $\bm v = [v_x, v_y]^T$ is the velocity in the EV frame. Besides, $\varphi$ is the heading angle referring to the global Cartesian frame, and $\omega$ is the yaw rate. The control input of the system is $\bm u = [a,\delta]^T$, where $a$ and $\delta$ represent the acceleration and steering angle, respectively. And we denote $m$ as the vehicle mass, $l_f$ and $l_r$ as the distance from the mass center to the front and rear axle, $k_f$ and $k_r$ as the cornering stiffness of the front and rear wheels, and $I_z$ as the inertia polar moment. For simplicity, we define $L_k=l_f k_f - l_r k_r$. Table \ref{param_model} lists the values of the above parameters. The model exhibits good numerical stability and high accuracy even when there are wide range speed changes as is typically observed in urban driving tasks~\cite{ge2021numerically}.
\begin{table}[h]
  \centering
  \caption{Parameter settings of the vehicle model}
  \resizebox{1.0\linewidth}{!}{
    \begin{tabular}{cccccc}
    \toprule
    Notation & Value & Unit & Notation & Value & Unit \\
    \midrule
    $k_f$    & -128916 & N/rad & $l_r$    & 1.85  & m \\
    $k_r$    & -85944 & N/rad & $m$     & 1412  & kg \\
    $l_f$    & 1.06  & m & $I_z$    & 1536.7 & $\text{kg}\cdot \text{m}^2$\\
    \bottomrule
    \end{tabular}%
    }
  \label{param_model}%
\end{table}%

\subsection{Proposed Autonomous Driving Framework}\label{subsec:ibpcFramework}

As shown in Fig.~\ref{fig:idmc_framework}, after a destination is assigned to the EV, the global path is planned by the A$^*$ algorithm based on the waypoints in the HD Map. To get the waypoints to be followed in MPC, a real-time trajectory is generated by spline interpolation based on the EV's current state.

In terms of environmental perception, there are various options, and we provide an alternative using sensor fusion. Next, we formulate the objective function with five indices. Notice that the SV collision avoidance and traffic rules compliance functions are achieved by applying our presented APF. With a variety of PFs, the proposed integrated behavior planning and motion control scheme can implicitly generate behavior planning commands (such as overtaking or following). Finally, the OCP is formulated in a receding horizon manner and solved to generate control commands for the EV.

 \subsection{Optimal Control Problem Formulation}
This section formulates the behavior planning and motion control scheme, which leverages the MPC to solve an optimal control problem (OCP) in the receding horizon manner. 
The cost function is formulated as 
\begin{align}
\label{basic_cost}
\begin{aligned}
    J(\bm x, \bm u, \bm p_\text{env}) &= \sum_{\tau=1}^N \| \boldsymbol x_{\text{ref},\tau}- \boldsymbol x_\tau\| ^2_{\bm{Q}} + \sum_{\tau=1}^N \|\boldsymbol u_{\tau}\|^2 _{\bm{R}} \\
    & +\sum_{\tau=2}^N \|\bm u_{\tau}-\bm u_{\tau-1}\|^2 _{\mathbf{R}_d}+F(\bm p_\text{env}, \bm x),
\end{aligned}
\end{align}
where $N$ is the length of the receding horizon, $\bm x_{\text{ref},\tau}\in \mathbb{R}^{6}$ is the reference trajectory at the time step $\tau$. In (\ref{basic_cost}), $\mathbf{Q}\in \mathbb{R}^{6\times 6}$, $\mathbf{R}\in \mathbb{R}^{2\times 2}$, $\mathbf{R}_d\in \mathbb{R}^{2\times 2}$ are positive semi-definite diagonal weighting matrices for penalizing reference tracking, energy saving, and passenger comfort, respectively. $F(\bm p_\text{env}, \bm x)$ is the APF for obstacle avoidance and traffic rule compliance. 
Therefore, the OCP is constructed as:
\begin{equation}
\label{NMPCOptProb}
\begin{array}{ll}
\underset{{\boldsymbol{x}}, {\boldsymbol{u}}}{\min} & J(\bm x, \bm u, \bm p_\text{env})\\
\text { s.t.} & \boldsymbol x_{\tau+1}=f(\boldsymbol x_\tau,\boldsymbol u_\tau), \forall \tau \in \{1,2,...,N\} \\
& -\boldsymbol u_\text{min} \preceq \boldsymbol u_\tau \preceq \boldsymbol u_\text{max}, \forall \tau \in \{1,2,...,N\} \\
& -\boldsymbol x_\text{min} \preceq \boldsymbol x_\tau \preceq \boldsymbol x_\text{max}, \forall \tau \in \{1,2,...,N\}.
\end{array}
\end{equation}
Note that the first constraint represents the vehicle dynamics (\ref{bicycle_dynamics_equation}). The second and third constraints indicate the restriction on the control inputs and states, which are based on the physics of the selected vehicle model. Problem (\ref{NMPCOptProb}) can be reformulated to a nonlinear programming (NLP) problem and solved by IPOPT in CasADi~\cite{Andersson2019}. Direct multiple shooting can be used, in which we treat the states at shooting nodes as decision variables and improves convergence by lifting the NLP problem to a higher dimension.  
\section{Traffic Participants and Traffic Rules Modelling} \label{PF design Sub_sec}
\begin{figure}[ht]
    \centering
    \includegraphics[width=1\linewidth]{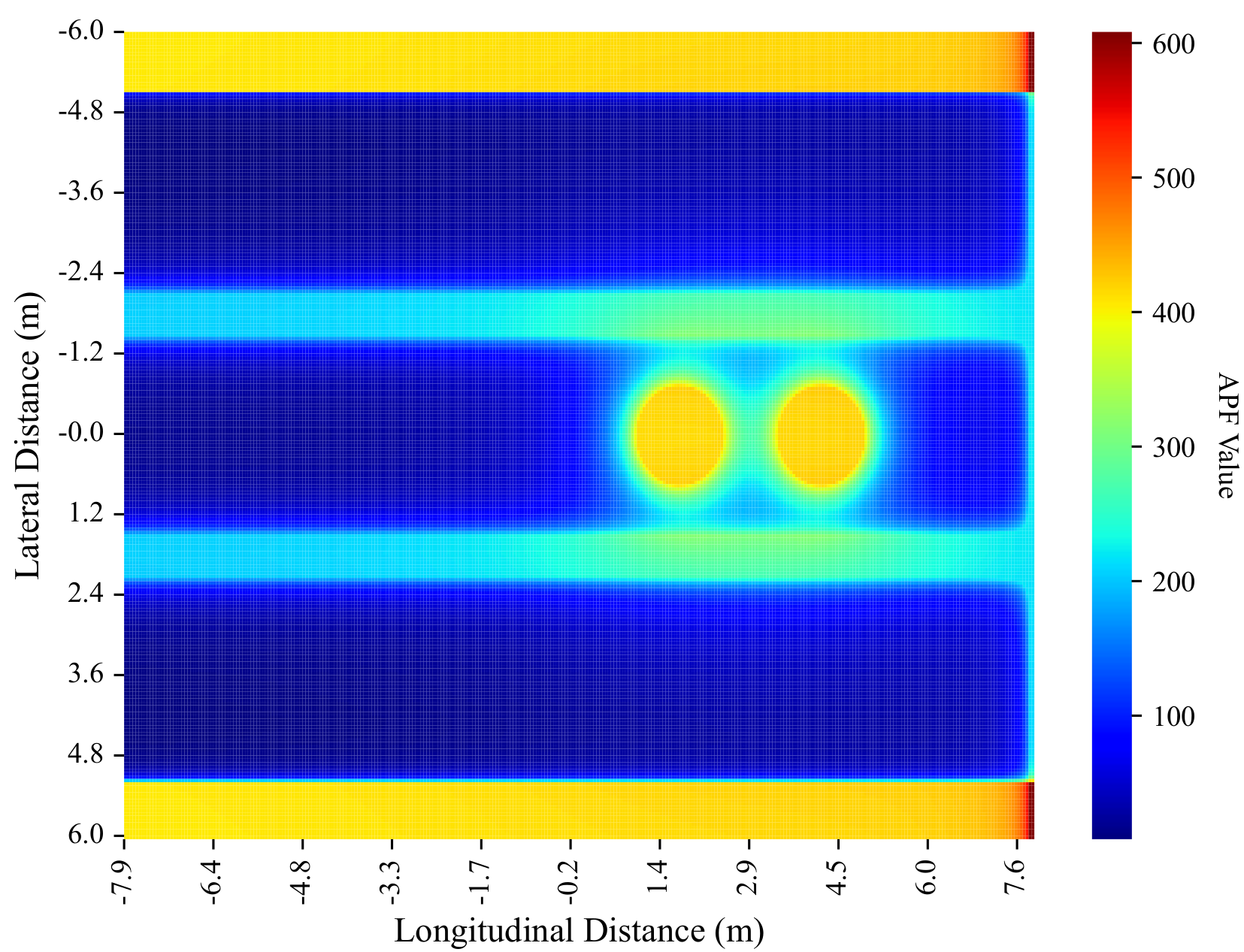}
    \caption{The APF is formed by the designed PFs, consisting of a three-lane road structure with two traversable ($F_\text{TR}$) and two non-traversable ($F_\text{NR}$) lane markings. The lateral positions of these markers are [5.25, -5.25] m and [1.75, -1.75] m, respectively. Additionally, the vehicle PF $F_\text{V}$ is created using two circles at the position of $[p_{kx}, p_{ky}]=[3.0,0]$ m. Lastly, the red traffic light PF $F_\text{TL}$ is at $p_y^\text{ev} = 7.9$\,m.}
    \label{fig:apf_bev}
\end{figure}
In this paper, the repulsive PF for surrounding vehicles, the repulsive PF for non-traversable lane markings, and the attractive PF for the centerline of lanes are designed. Unlike the traditional APF method, where the target position is set as a gravitational field~\cite{weerakoon2015artificial}, this paper employs several novel penalty terms to form the potential field in the integrated behavior planning and control module:
\begin{equation}
\label{eq:sum_PFs}
    F(\bm p_\text{env}, \bm x) = \sum_{K\in \mathcal K} F_K = \sum_{K\in \mathcal K} \sum_{i\in \mathcal{I}_K} f_K(\bm p_K[i],\bm x),
\end{equation}
where each element of the potential field $F_{K}$ is constructed by all the PFs of the detected specific objects in class $k$. And $\bm p_\text{env}$ is concatenated by $\bm p_K$, $\mathcal K := \{\text{NR}, \text{TR}, \text{V}, \text{TL}\}$, which elements denote non-traversable lane markings, traversable lane markings, surrounding vehicles, and dynamic traffic lights, respectively. $\mathcal{I}_K:=\{1,2,...,n_K\}$ denotes the objects in class $K$. Noted, the notation of $f_K(\bm p_K[i],\bm x)$ is simplified as $f_{K
,i}$ in the following parts. To simplify the PF design, we classify existing lane markings into two categories: traversable lane markings represented by broken lines and non-traversable lane markings represented by solid lines. 

The lateral distance from the EV to the lane marking is an essential value to determine the PF force. We define the road centerline state vector as $\bm p_i = [p_{xi}, p_{yi}, \beta_i]^T$, where the first two items $p_{xi}$ and $p_{yi}$ represent the position of the current road centerline in global Cartesian frame, and the last item $\beta_{i}$ is the tangential angle of the current road. Therefore, the lateral position of the lane marking is given as
\begin{align} \label{cordinate transform of rd}
s_{\text{R}}(\bm x, \bm p_i)=\left\{
\begin{aligned} 
&\left({p}_y^\text{ev}(\bm p_i) + \frac{w_R}{2}\right) - y^\text{ev}(\bm x, \bm p_i) &\text{left}\\ 
&y^\text{ev}(\bm x, \bm p_i) - \left(p_y^\text{ev}(\bm p_i) - \frac{w_R}{2}\right)
&\text{right}
\end{aligned}
\right.  
\end{align}
where left and right indicate the lane marking on the left or right side of the vehicle, $w_R$ is the lane width, $p_y^\text{ev}$ and $y^\text{ev}$
\begin{align}
&p_y^\text{ev}(\bm p_i) = p_{xi} \sin(\beta_i) + p_{yi}\cos(\beta_i),\\
    \begin{split}
        &y^\text{ev}(\bm x,\bm p_i) = p_{x} \sin(\beta_i) + p_{y}\cos(\beta_i),
    \end{split}
\end{align}
 are the lateral positions of the lane centerline and the EV in the heading angle-paralleled (HAP) frame, where the orientation of the road is parallelled with the $x$-axis.
\subsection{Non-Traversable Lane Markings} 
To comply with traffic rules, vehicles must not traverse a solid lane marking. Therefore, when the distance between the EV and the surrounding lane markings decreases to a certain extent, the designed PF should give a repulsive force that increases rapidly with decreasing distance to urge the EV to steer away immediately. Based on the inverse proportional function and power function, we designed the following PF for the $i$th non-traversable lane marking:
\begin{align} 
f_{\text{NR},i}=\left\{
\begin{aligned} 
&m_s&s_{\text{R}_i}\leq0.1\\ 
&\frac{a_\text{NR}}{s_{\text{R}}(\bm x, \bm p_\text{NR}[i])^{b_\text{NR}}} - e_s& 0.1<s_{\text{R}_i}<1.5\\
&0& s_{\text{R}_i} \geq 1.5
\end{aligned}
\right. ,
\end{align}
where $a_\text{NR}$ and $b_\text{NR}$ are the intensity and shape parameters that control the strength of the repulsive force and the effective range of the PF. The smooth parameters
$$e_s = \frac{a_\text{NR}}{1.5^{b_\text{NR}}},\ m_s = \frac{a_\text{NR}}{0.1^{b_\text{NR}}}-e_s$$ guarantee the continuity of the PF value when $s_{\text{R}_i}$ switches between conditions. The heat map of the designed non-traversable lane marking's PF is illustrated on both sides of the lateral direction in Fig. \ref{fig:apf_bev}, where $a_\text{NR}=100$ and $b_\text{NR}=2.0$.
\subsection{Traversable Lane Markings}
In the process of overtaking and lane changing, it is necessary to traverse the broken lane marking on the road. 
When the reference lane of the EV is blocked from ahead by other traffic participants, the EV needs to switch to an adjacent lane and keeping driving until the reference lane is free.
When EV is driving on a lane that is not the reference one, designing a PF that keeps the EV driving along the centerline is necessary, as otherwise the EV will be attracted to the side of the current lane due to the attractive force of the reference lane.
Thus, the PF of the $j$th traversable lane marking is designed as follows:
\begin{align} \label{CR-PF}
f_{\text{TR},j}=\left\{
\begin{aligned} 
&a_\text{TR} (s_{\text{R}}(\bm x, \bm p_\text{TR}[j])-b_\text{TR})^2& s_{\text{R}_j}<b_\text{TR}\\
&0& s_{\text{R}_j} \geq b_\text{TR}
\end{aligned}
\right. ,
\end{align}
where $a_\text{TR}$ is the intensity parameter and $b_\text{TR}$ is the effective range of the repulsive force from the lane marking. Such PF repulses the EV be away from the lane marking and stay around the lane center. The heat map of the aggregated $F_\text{TR}$ is shown in the middle area of the lateral direction in Fig. \ref{fig:apf_bev}, where the parameters are $a_\text{TR}=20$ and $b_\text{TR}=1.0$.


\subsection{Surrounding Vehicles}
The main traffic participants on the road are vehicles and pedestrians, both of which are strictly prohibited to collide. 
We approximate each on-road vehicle with two identical circles aligned in the longitudinal direction that covers the entire vehicle.
In order to avoid surrounding vehicles smoothly and quickly, we design the following PF:
\begin{equation}\label{Vehicle_PF}
    f_{\text{V},k} = \sum _{q=1}^2\sum_{o=1}^2 \frac{a_\text{V}}{((p_{xq}-p_{kox})^2+(p_{yq}-p_{koy})^2)^{b_\text{V}}},
\end{equation}
where $a_\text{V}$ and $b_\text{V}$ are the intensity and shape parameters, which control the strength of the repulsive force and the effective range of this PF. As the vehicle is represented by two circles, we must calculate the circles' positions and sum their force to formulate the integrated PF. The position vector $[p_{kox},p_{koy}]^T$ denotes position of the $o$th circle of the $k$th surrounding vehicle, and we have
\begin{equation}
\left[
\begin{array}{c}
p_{kox} \\
p_{koy}
\end{array}
\right]=
\left[
\begin{array}{cc}
1& 0 \\
0& 1
\end{array}
\right]
\left[
\begin{array}{c}
p_{kx} \\
p_{ky}
\end{array}
\right]
\pm
\left[
\begin{array}{c}
\cos{\beta_i}\\
\sin{\beta_i}
\end{array}
\right] r_\text{V},
\end{equation}
where the common radius of the circles is $r_\text{V}=1.2$ and $o = \{0,1\}$ indicates whether the circle is front or rear.
Fig. \ref{fig:apf_bev} illustrates the potential field of a SV. And the parameters are set as $a_V = 5.0$, and $r_V = 1.2$ m.

\subsection{Traffic Control Signals}
In addition to obeying the rules of lane markings, vehicles must also obey the command of traffic lights. This paper proposes a novel PF for traffic lights, which can restrict the lateral and longitudinal positions of the EV simultaneously. The PF applies repulsive forces in three directions to the ego vehicle without changing its original reference trajectory. In particular, the PF in front of the EV is designed as
\begin{equation}
\label{TL_equation}
    f_{\text{TL},1} = c_\text{TL}\left(\frac{a_{\text{TL}_1}}{d_{x}^\text{ev}}+\frac{a_{\text{TL}_2}}{d_{yl}^\text{ev}}+\frac{a_{\text{TL}_2}}{d_{yr}^\text{ev}}\right),
\end{equation}
where $c_\text{TL}$ is an indicator of the state of the traffic light, which equals to 0 is the traffic light is $green$ and 1.0 otherwise.
Besides, $a_{\text{TL}_1}$ and $a_{\text{TL}_2}$ are intensity parameters in the longitudinal and lateral orientations, respectively, and ${d_{x}^\text{ev}}$ is the longitudinal distance from EV to the stop line. Moreover, ${d_{yl}^\text{ev}}$, $d_{yr}^\text{ev}$ are the distance of the EV to the left and right lane markings, respectively, which can be calculated by (\ref{cordinate transform of rd}) as well. As shown in the right side of Fig.~\ref{fig:apf_bev}, the $F_\text{TL}$ (with $a_{\text{TL}_1}=20$ and $a_{\text{TL}_2}=40$) is able to urge the EV to stay behind the stop line when the traffic light is red.

\section{Simulation Results}
\begin{figure}[htbp]
\centering
\subfigure[Roundabout]{
\includegraphics[width=1\linewidth]{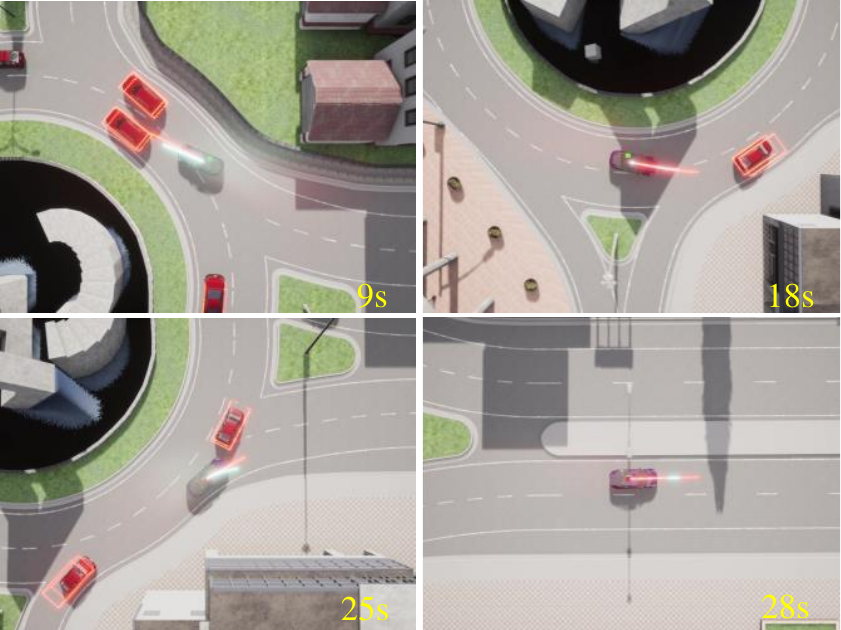}
\label{fig:scenario_roundabout}
}
\subfigure[Multi-lane Adaptive Cruise Control]{
\includegraphics[width=1\linewidth]{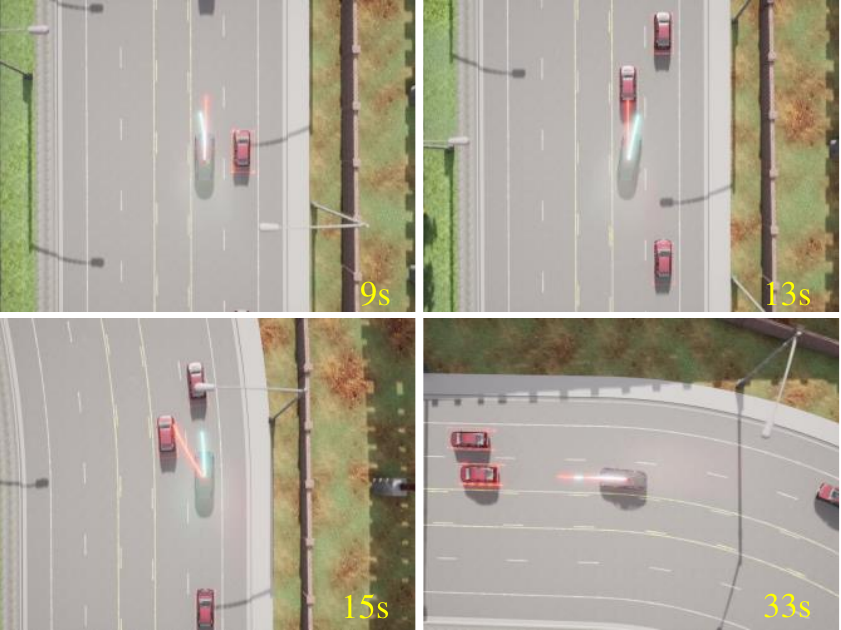}\label{fig:scenario_ACC}
}
\subfigure[Crossroad]{
\includegraphics[width=1\linewidth]{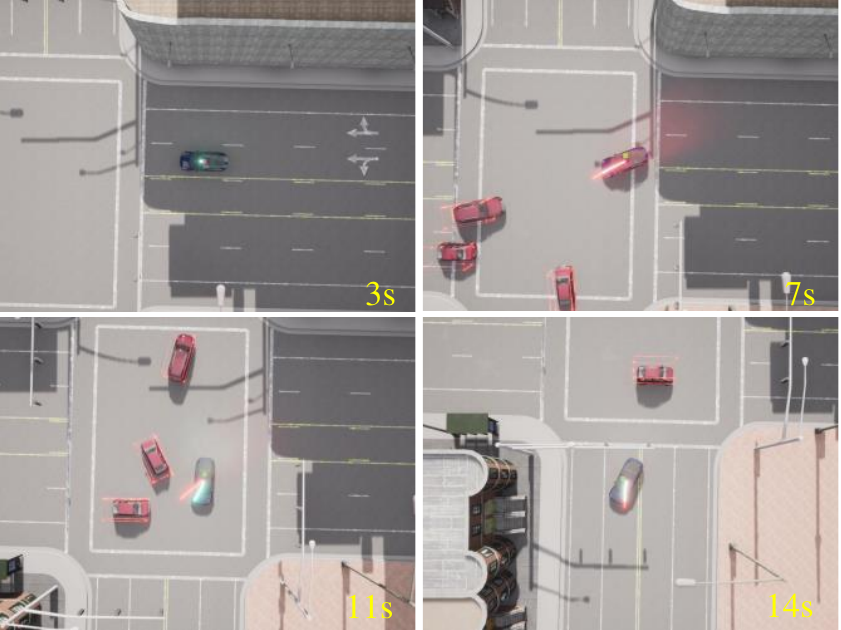}                  
\label{fig:scenario_crossroad}
}
\caption{Simulation on three illustrative urban driving scenarios in CARLA. Four key frames are selected for each scenario, and the velocity reference is set to 35\,km/h, 40\,km/h, 20\,km/h, respectively, for the three scenarios from left to right. The red line on the EV represents the trajectory generated from the spline interpolation described in Section~\ref{subsec:ibpcFramework} while the cyan line shows the optimized predicted trajectory with receding horizon $N=10$.}
\label{fig:scenarios}
\end{figure}
In this section, we comprehensively evaluate the proposed framework towards three different representative urban driving scenarios. 
We implement the proposed scheme in CARLA~\cite{dosovitskiy2017carla} based on Ubuntu 20.04 with AMD Ryzen 5600G CPU and NVIDIA RTX 3060 GPU. 
\subsection{Simulation Scenarios}
\subsubsection{{Scenario 1: Roundabout Driving}}
In the roundabout scenario, the EV  is going to enter the roundabout from the east entrance and leave at the opposite exit with a reference velocity $v_\text{ref}=35$\,km/h. Typically, the heavy traffic flow renders great difficulties for this task. Moreover, the frequent and unpredictable lane-changing behavior of other vehicles makes the driving task even more challenging.
\subsubsection{Scenario 2: Multi-Lane Adaptive Cruise Control}
In this scenario, the EV aims to follow the current lane at a given speed $v_\text{ref}=40$\,km/h while keeping a safe distance from the leading vehicle until reaching the terminal state. It is pertinent to note that, apart from maintaining a safe distance from surrounding vehicles, traversable and non-traversable lane markings should also be considered in this task.

\subsubsection{{Scenario 3: Crossroad Driving}}
To get through the crossroad safely, the EV, with $v_\text{ref}=20$\,km/h, needs to comply with the state change of the traffic light. In this task, the global path leads the vehicle to a left turn at the dense crossroad where the traffic light is about to turn red as the EV approaches. Meanwhile, the non-traversable lane markings constraints cannot be violated during the turning process. 


\subsection{Overall Performance}\label{subsection:OverviewPerformance}
\begin{figure*}[htbp]

\centering
\subfigure[Tracking error in Scenario 1]{
\includegraphics[ width=0.31\linewidth]{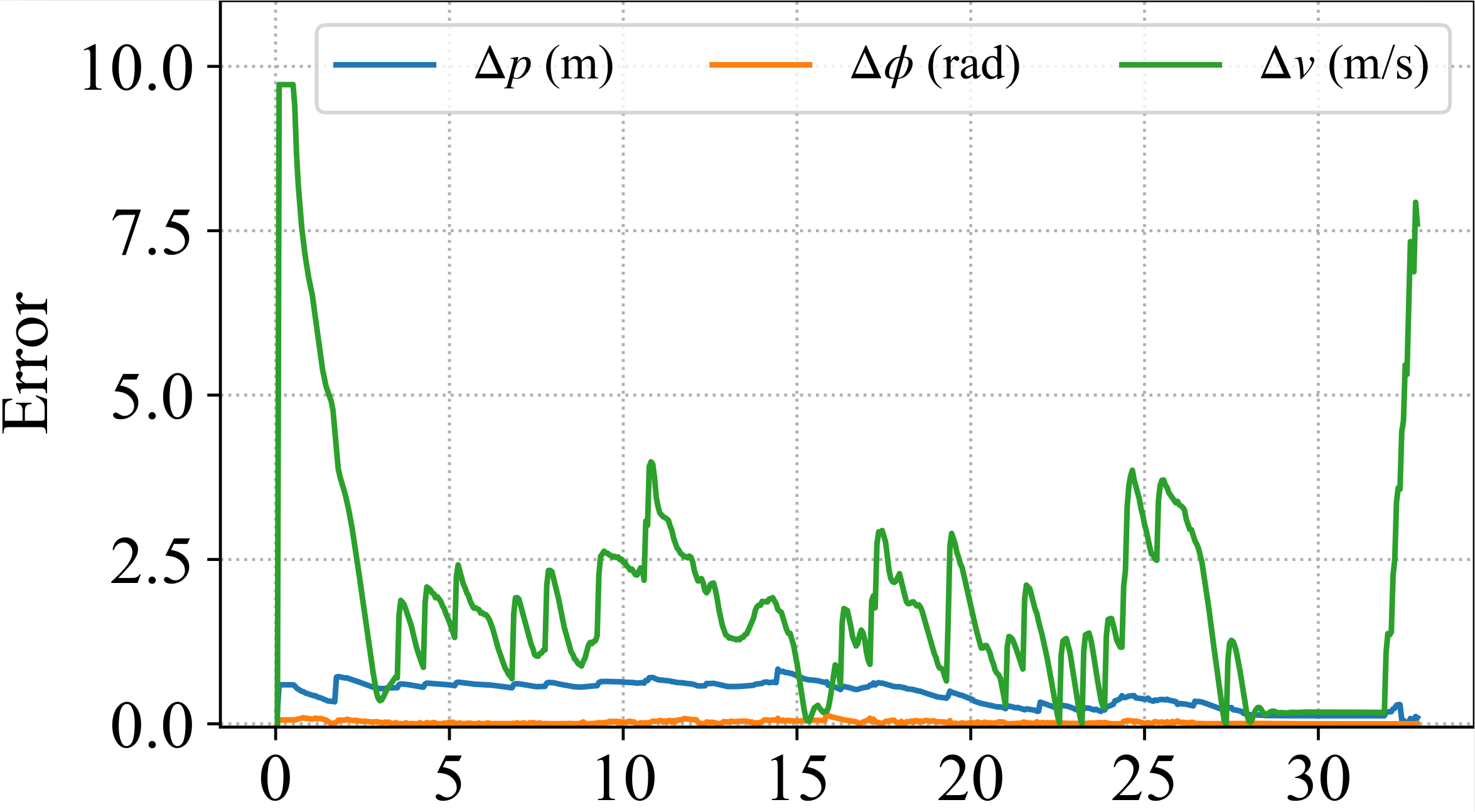}
\label{fig:error_s1}
}
\subfigure[Tracking error in Scenario 2]{
\includegraphics[width=0.31\linewidth]{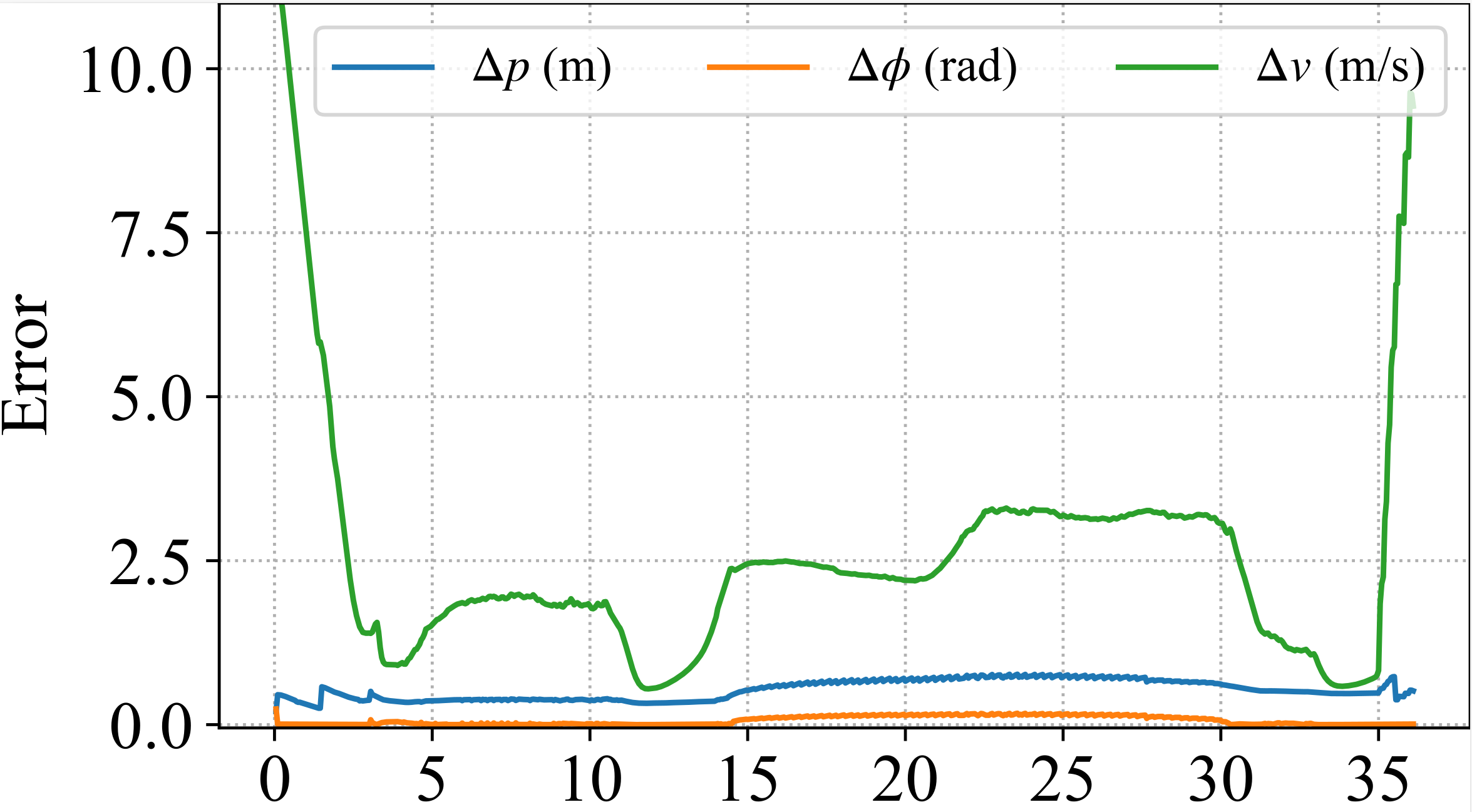}
\label{fig:error_s2}
}
\subfigure[Tracking error in Scenario 3]{
\includegraphics[width=0.31\linewidth]{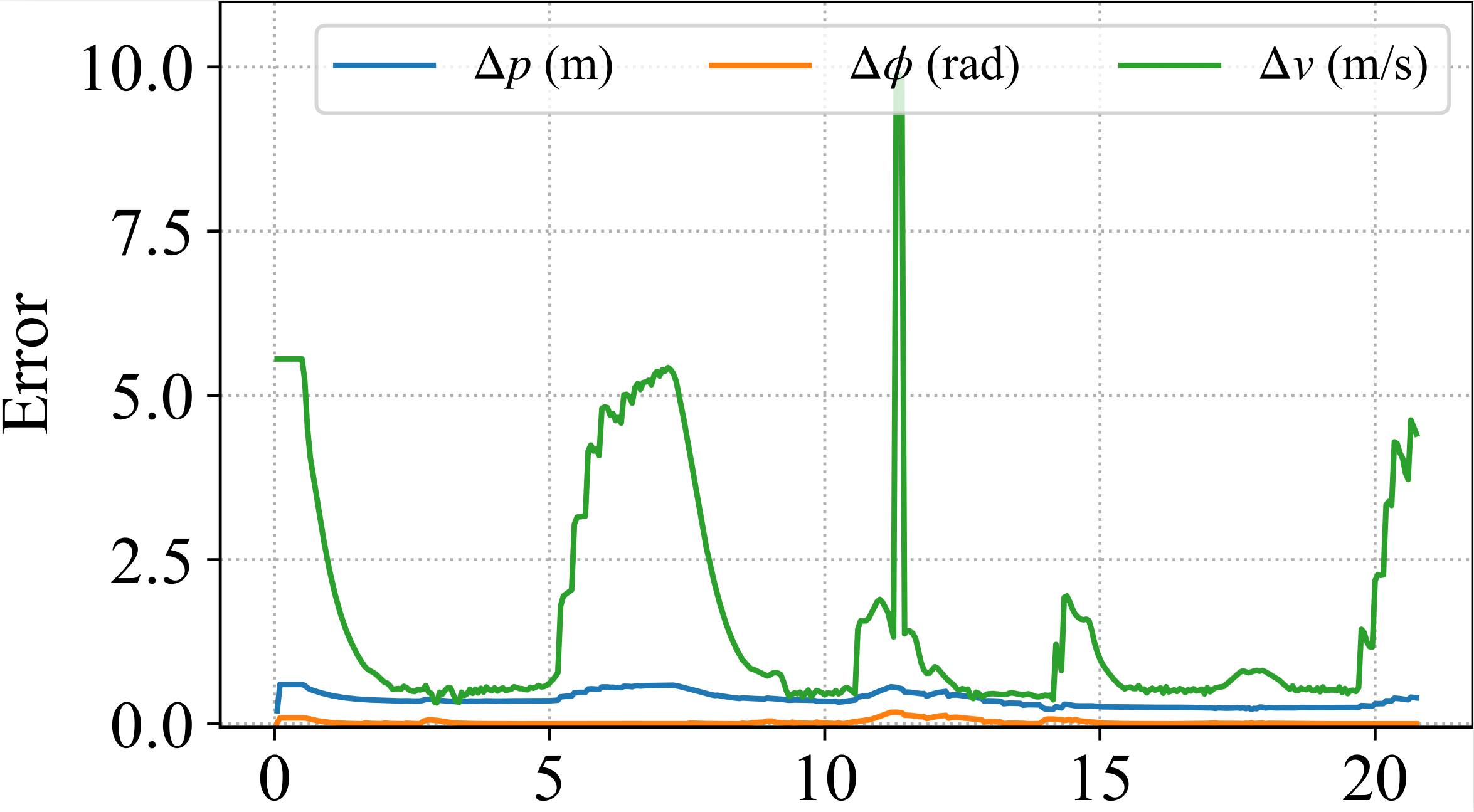}
\label{fig:error_s3}
}
\subfigure[PF value in Scenario 1]{
\includegraphics[width=0.31\linewidth]{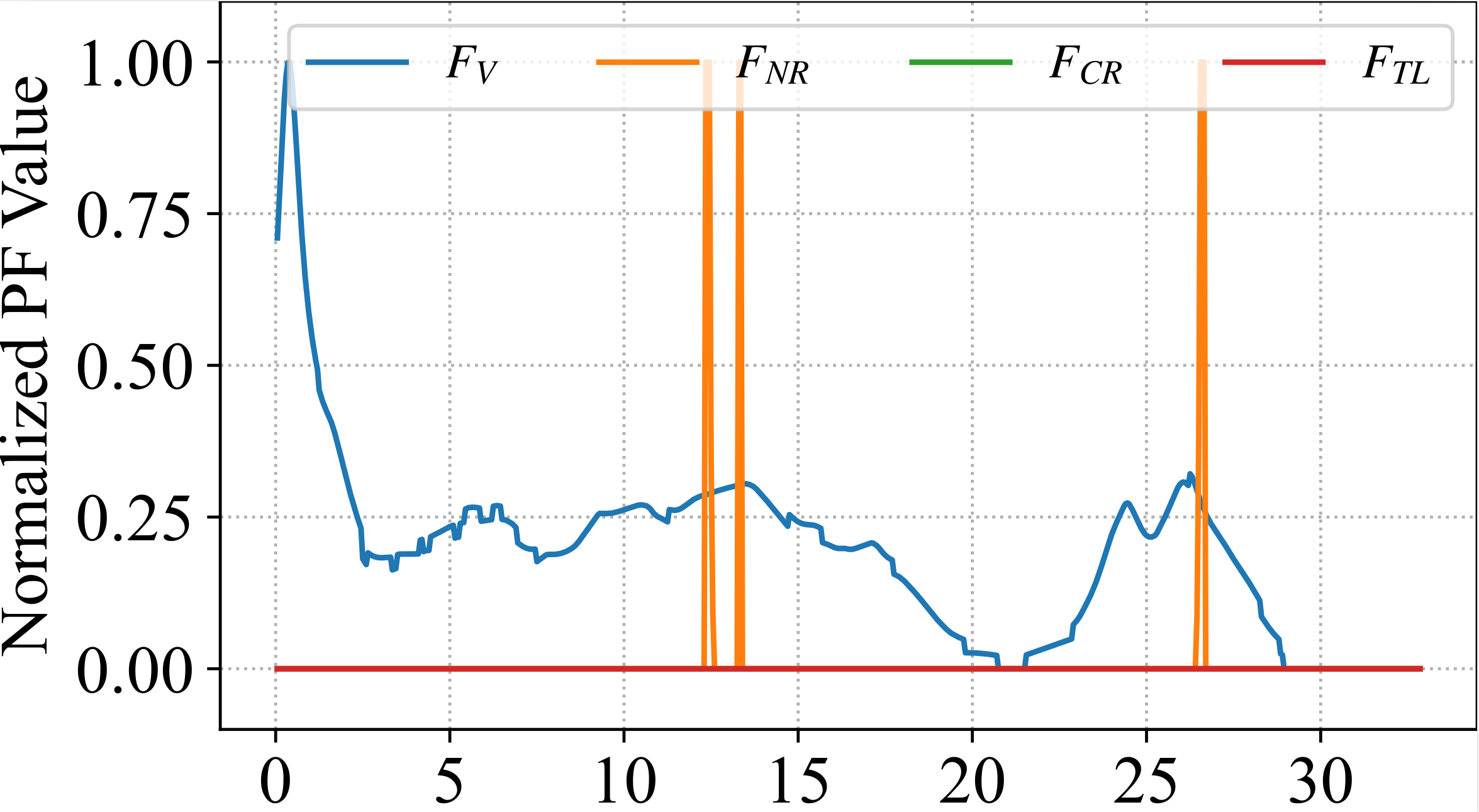}
\label{fig:r_apf}
}
\subfigure[PF value in Scenario 2]{
\includegraphics[width=0.31\linewidth]{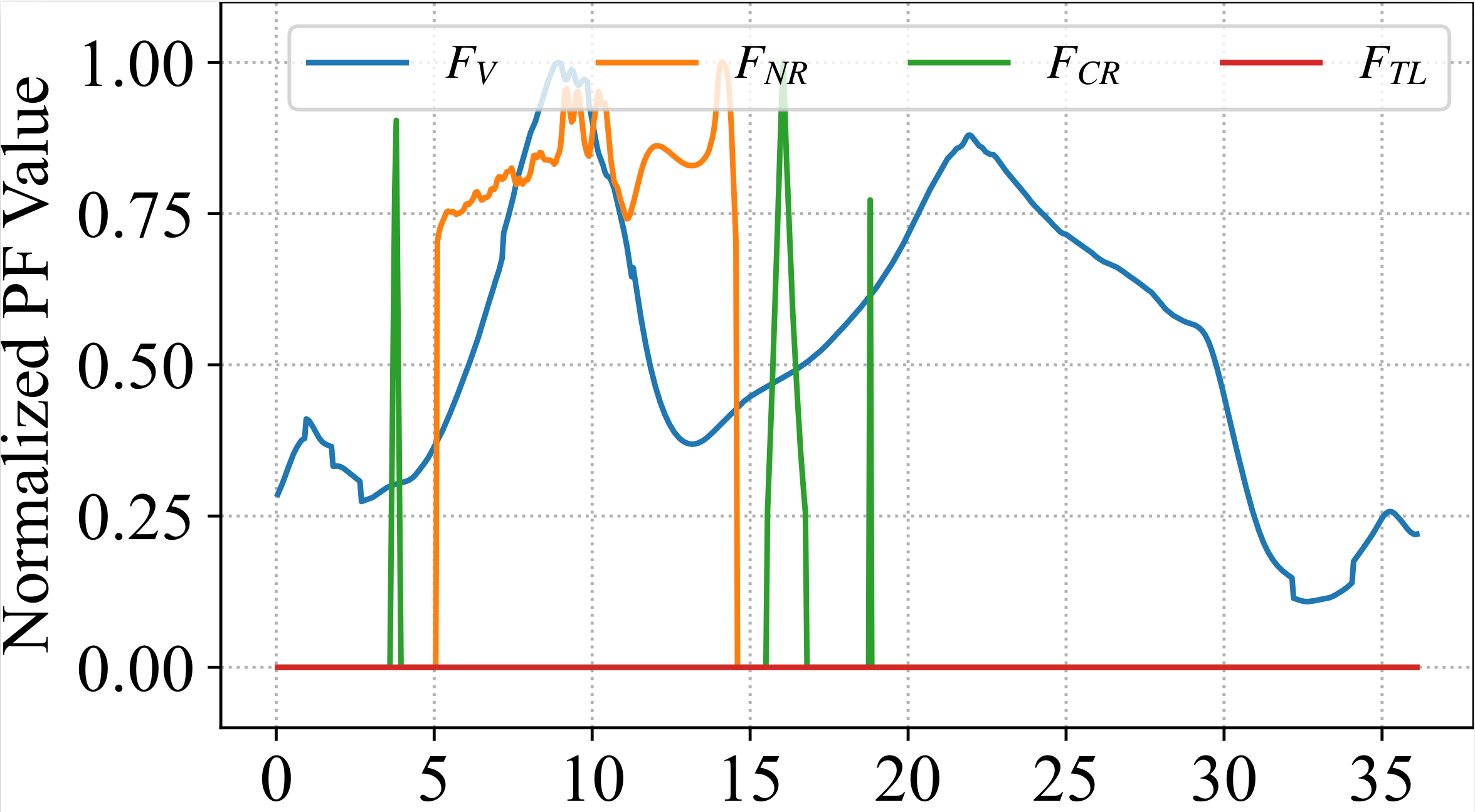}
\label{fig:acc_apf}
}
\subfigure[PF value in Scenario 3]{
\includegraphics[width=0.31\linewidth]{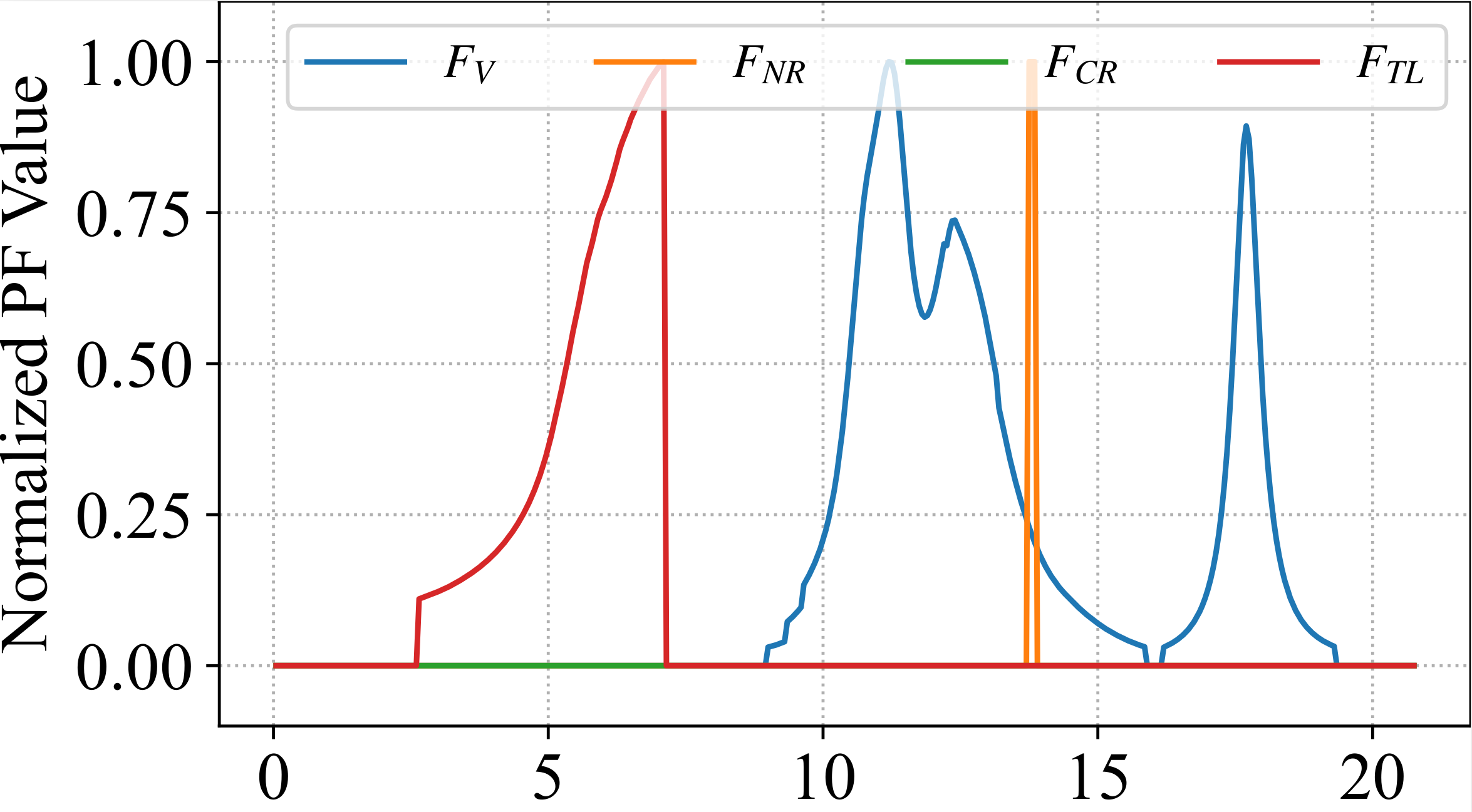}
\label{fig:c_apf}
}

\subfigure[Control input in Scenario 1]{
\hspace{0.1cm}\includegraphics[width=0.31\linewidth]{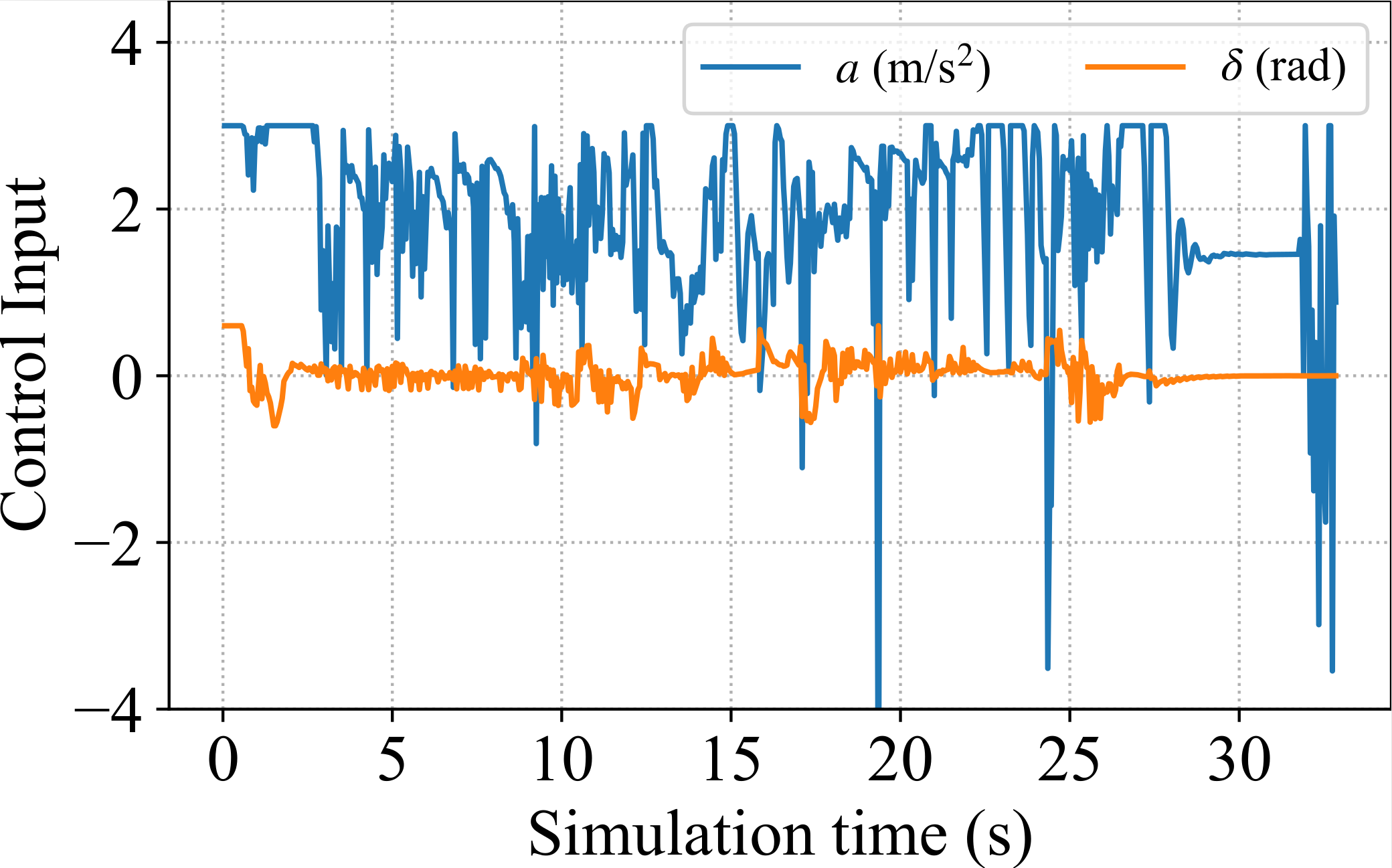}
\label{fig:ctrlInput_s1}
}
\subfigure[Control input in Scenario 2]{
\includegraphics[width=0.31\linewidth]{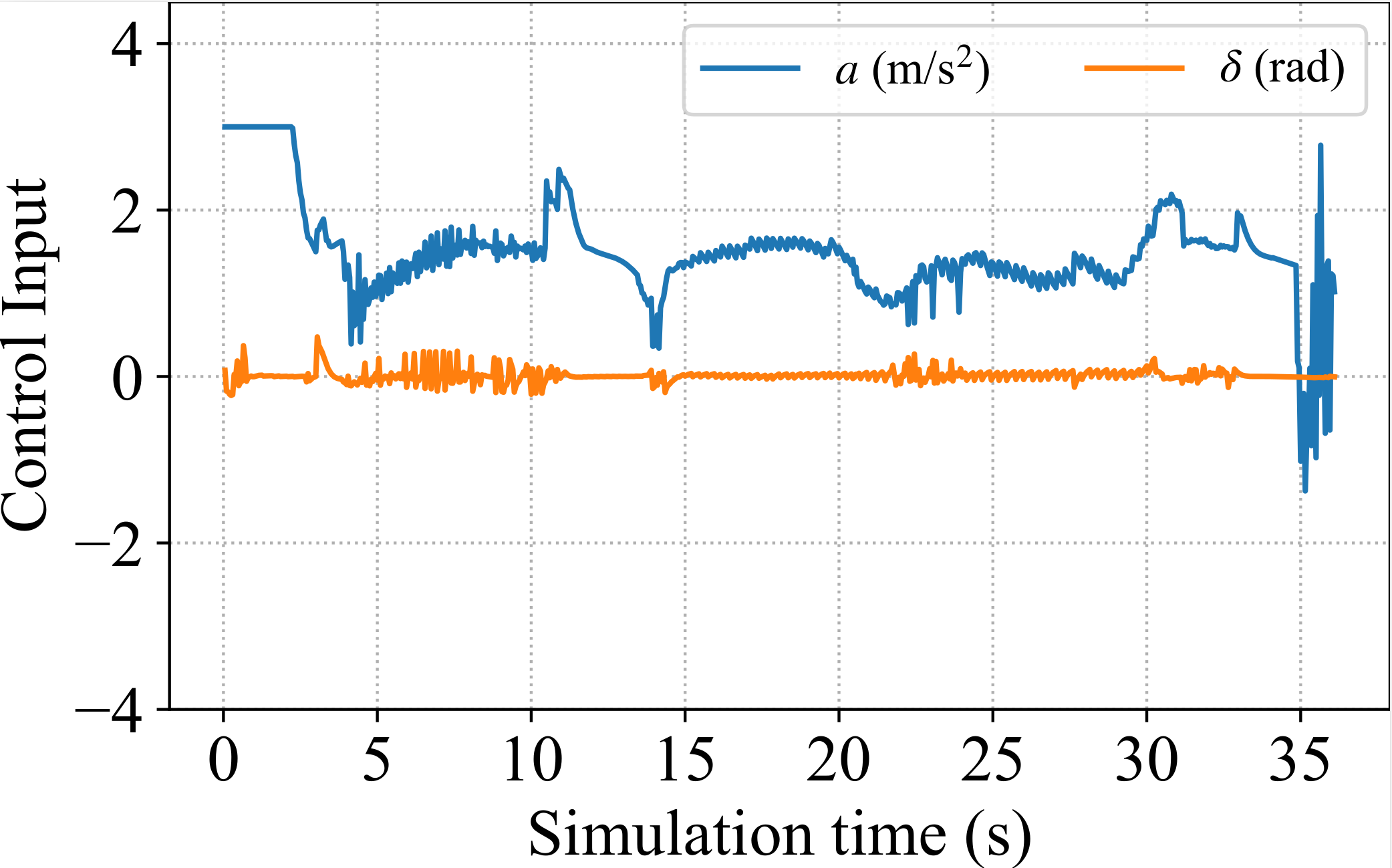}
\label{fig:ctrlInput_s2}
}
\subfigure[Control input in Scenario 3]{
\includegraphics[width=0.31\linewidth]{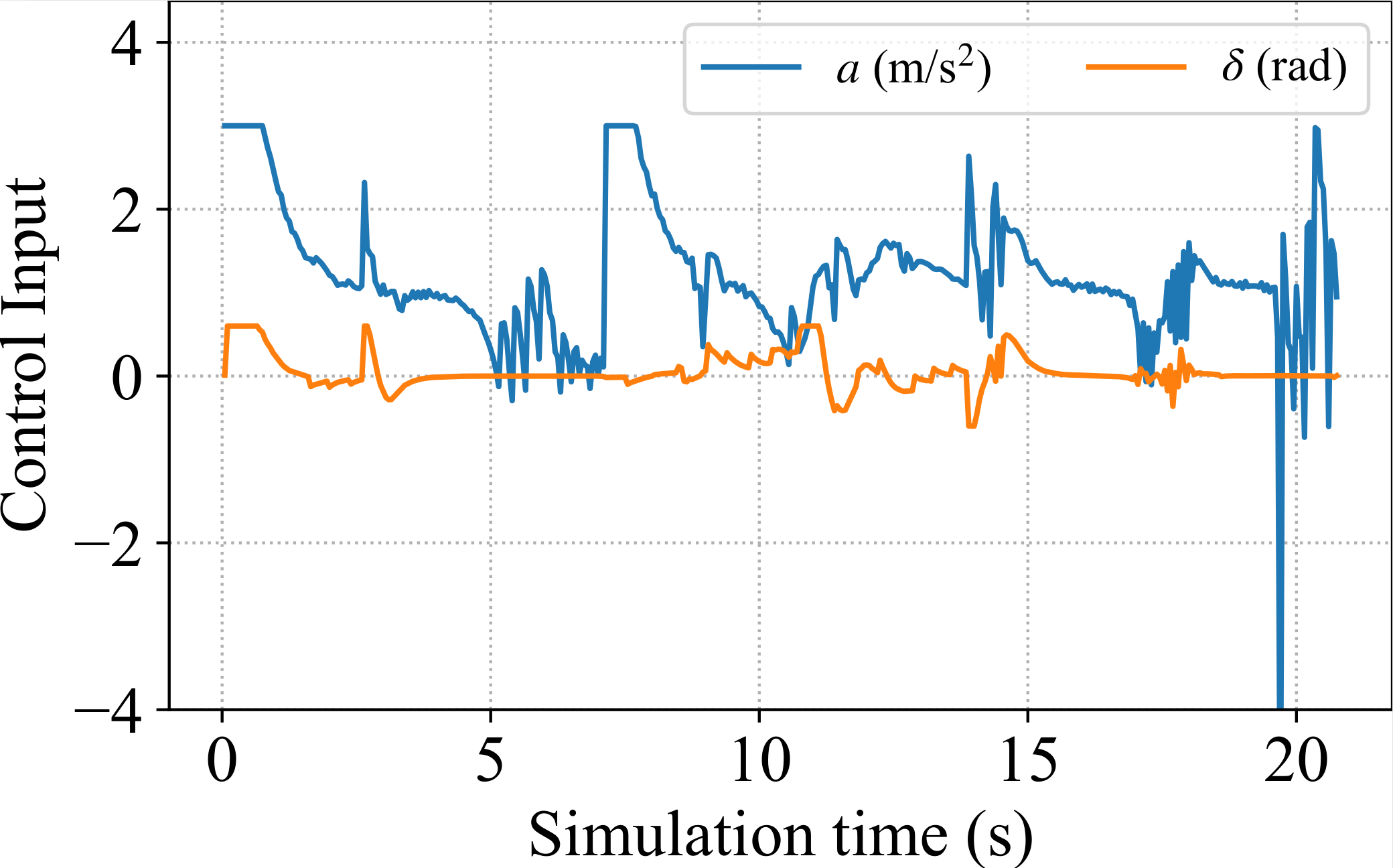}
\label{fig:ctrlInput_s3}
}
\DeclareGraphicsExtensions.
\caption{Tracking error, PF value, and control input from one trail in three urban driving scenarios. 
(a), (b), (c) record the position error $\Delta p$, steering angle error $\Delta \varphi$, and target velocity tracking error; (d), (e), (f) record the PF values including $\sum F_\text{V}$, $\sum F_\text{NR}$, $\sum F_\text{TR}$, and $\sum F_\text{TL}$; (g), (h), (i)  record the acceleration $a$ and steering angle $\delta$.}
\label{fig:allStaticsSimulation}
\end{figure*}

In the implementation of the proposed integrated behavior planning and motion control framework, the prediction horizon is set as $N = 10$ and the sampling time is set as $\Delta T = 0.05$\,s. In each scenario, we select four key frames for visualization from one trial shown in Fig. \ref{fig:scenarios}, with their tracking error shown in Fig. \ref{fig:allStaticsSimulation}(a)-(c) respectively. 
In Scenario 1, as shown in Fig. \ref{fig:scenario_roundabout}, the EV is commanded to change its lane to enter/exit the roundabout at $9$\,s and $18$\,s. However, the planned path is blocked by the vehicles in front of it. Thus, the EV keeps finding an appropriate chance to change lane, which makes the curve of $\Delta v$ fluctuates violently. 
For Scenario 2, as shown in Fig. \ref{fig:scenario_ACC}, the EV is desired to maintain a speed at 40\,km/h. In this trial, the leading vehicle slows down at $9$\,s, so that the EV merges into its neighbor lane temporarily at $13$\,s in order to maintain the desired forward velocity; and return to the original lane at $15$\,s to lower the tracking error.
In terms of Scenario 3, at the crossroad, the reference speed for the EV is set as 20\,km/h. As shown in Fig. \ref{fig:scenario_crossroad}, the EV stops smoothly at the crossroad after the traffic light turns red at $3$\,s, which leads to the gradually increased velocity error around $t = 5$\,s. The peak velocity error at  $11$\,s is caused by the sudden acceleration of the two surrounding vehicles at the intersection.     
To further demonstrate the effectiveness of our proposed integrated behavior planning and motion control scheme, we execute the simulation of 30 trials for each scenario and evaluate the performance quantitatively. We define a trial without any vehicle collision as a success. The success rate, computation time together with the average tracking error are listed in Table \ref{tab:overall}. 
\begin{table}[h]
  \centering
  \caption{Overall Performance in Dynamic Urban Driving Simulation}
  \resizebox{1.0\linewidth}{!}{
    \begin{tabular}{ccccccc}
    
    \toprule
    \multirow{2}[1]{*}{Scenarios} & \multirow{2}[1]{*}{Succ. Rate (\%)} & \multicolumn{3}{c}{Average Tracking Error} & \multicolumn{2}{c}{Computation Time (ms)} \\
\cmidrule{3-7}          &       & $\Delta p $\,(m)     & $\Delta v$\,(m/s)     & $\Delta \varphi$\,(rad)   & Ave.$\pm$SD & Max \\
    \midrule
    Roundabout & \textbf{93.3}  & 0.495 & 2.340 & \textbf{0.020} & 9.39$\pm$3.40 & \textbf{24.09} \\
    Multi-lane ACC   & 84.0    & 0.468 & 1.742 & 0.031 & 11.00$\pm$3.41 & 48.40 \\
    Crossroad & 80.0    & \textbf{0.424} & \textbf{1.640} & 0.022 & \textbf{6.86$\pm$2.38} & 30.08 \\
    \bottomrule
    \end{tabular}%
    }
  \label{tab:overall}%
\end{table}%
Therefore, our proposed method can be applied to the three highly dynamic scenarios with a satisfactory success rate and good tracking performance. In particular, we achieve the highest success rate in the roundabout scenario at 93.3\%. As for the tracking performance, the velocity term is the most significant tracking error among the three scenarios due to the difficulty of maintaining a constant speed in the highly dynamic environment with unpredictable traffic elements, which is consistent with our analysis before on one trail from Fig.\,\ref{fig:error_s1}. Yet we can still attain a velocity error lower than 2.3404\,m/s in the challenging scenarios. Moreover, the formulated OCP can be solved within milliseconds, from which the real-time performance of the integrated behavior planning and motion control scheme is demonstrated.         

With the above analysis, we conclude that the proposed urban autonomous driving framework is able to generate safe reactions,  and keep high traffic efficiency with high-level behavior planning in real-time with flexible maneuvers.



\subsection{Effectiveness Analysis of the Designed APF}
In this section, the effectiveness of the embodied APF in the proposed integrated framework is evaluated. Observed from Fig. \ref{fig:allStaticsSimulation}(d)-(f), the four PFs effected very well. Their combined effect allows the EV to drive reactively, smoothly, and safely to its destination in all three scenarios in Fig. \ref{fig:scenarios}.
To ensure traffic safety and avoid collision with other vehicles, the vehicle PF should emit repulsive force to the EV when it gets closer to the surrounding ones. But if a surrounding vehicle gets too close, it could generate a relatively large repulsive force on the EV, and the EV will diverge from the target lane. So exerting a repulsive force from the lane marking is necessary to urge the EV to drive in its own lane, so that the traffic rule would be obeyed by not crossing non-traversable lane markings.  In Fig. \ref{fig:scenario_roundabout}, when the EV is exiting the roundabout before 25\,s, a surrounding vehicle is driving very closely, and it doesn't mean to exit the roundabout. At this moment, the repulsive force $F_\textup{V}$ is increasing and urges the EV to leave that vehicle. But as shown in Fig. \ref{fig:r_apf}, when EV gets closer to the road boundary, there exists a non-traversable lane marking, the repulsive force $F_\textup{NR}$ gets larger rapidly and urged the control output $\delta$ keep around 0, which avoids the EV from traversing the road boundary. Thus, it ensures the safety of the EV and prevents traffic rule violations.

Besides, the EV needs to keep in the middle of a strange lane, when it is urged to change lanes caused by some vehicles ahead occupying its original lane, like the situation shown in Fig.\,\ref{fig:scenario_ACC} around 13\,s. 
In the multi-lane adaptive cruise control (ACC) simulation, the EV is cruising on the highway at a given lane but the vehicle ahead blocks its way when $t=13$\,s. Because of the large $F_\textup{V}$, the EV changes to its right lane and stays in the middle autonomously with the help of the increasing $F_\text{TR}$ shown in Fig. \ref{fig:acc_apf}.

To obey the rules regarding dynamic traffic lights in the crossroad scenario in Fig. \ref{fig:scenario_crossroad}, the repulsive $F_\text{TL}$ increased smoothly but still quickly when the traffic light is getting ``red'', causing the EV to stop smoothly as shown in Fig.~\ref{fig:c_apf} so that provide a comfortable ride experience. Fig.\,\ref{fig:ctrlInput_s3} illustrates that when the traffic light becomes red from 3\,s to 7\,s, the acceleration $a$ became relatively large to brake the EV before the stop line.
In summary, the PF takes effective and urges the control input change reactively with PF values, when different urban driving situations are encountered.

\subsection{Failure Case Analysis}
To improve the success rate and enhance the urban driving safety of our proposed integrated behavior planning and motion control scheme, the failure cases in Section \ref{subsection:OverviewPerformance} are analyzed in detail. Collision failure occurs most frequently in the crossroad scenario. Traffic rule violation includes driving to a solid line marking and running a red light, which occurs in the roundabout and multi-lane ACC.
\begin{figure}[h]
    \centering
    \subfigure[Aggressive overtaking]{
    \includegraphics[width=0.46\linewidth]{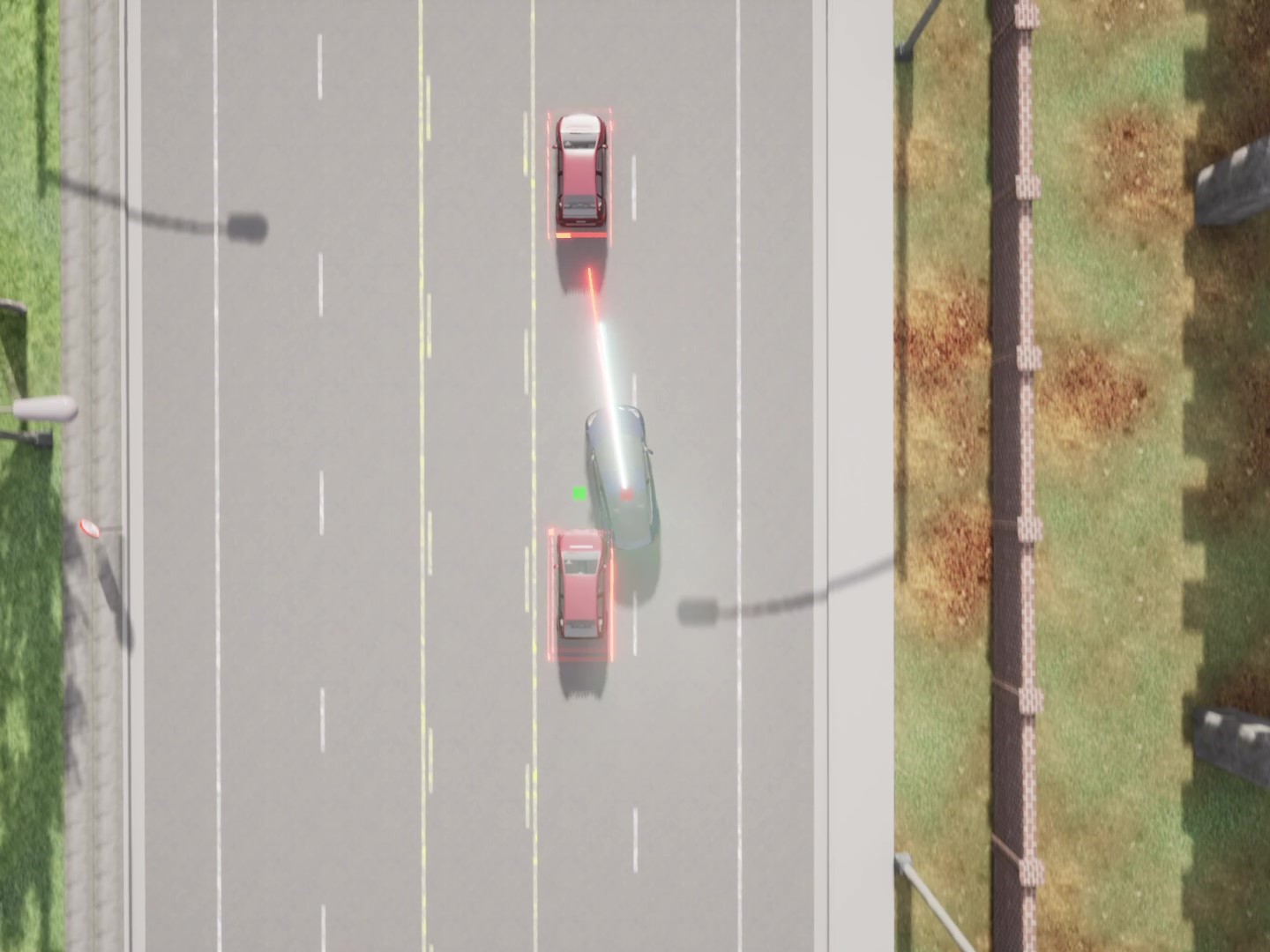}
    \label{fig:aggressiveOvertaking}
    }
    \subfigure[Driving to a solid line]{
    \includegraphics[width=0.46\linewidth]{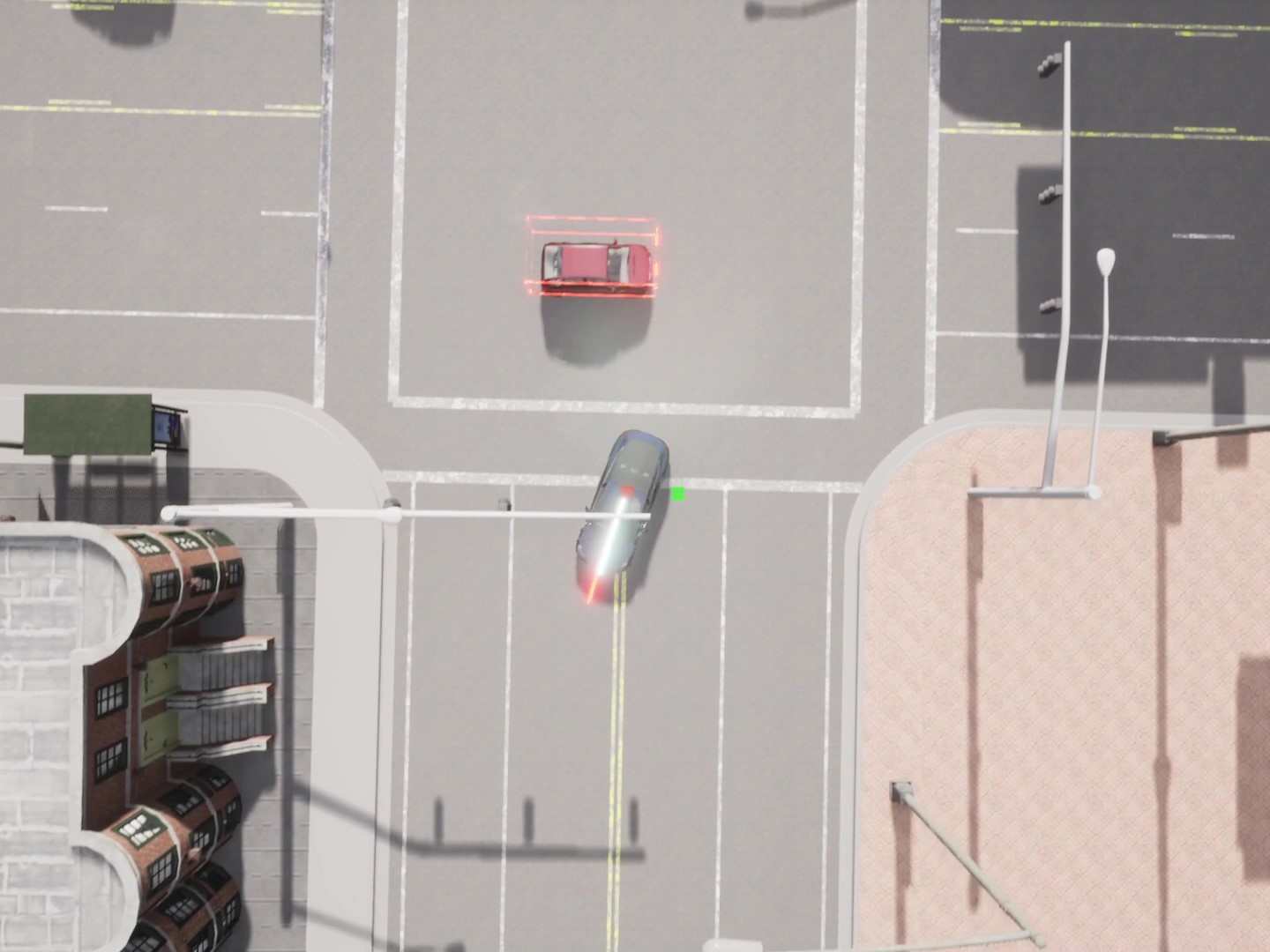}
    \label{fig:DrivingSolidLine}
    }
    \caption{Typical failure cases in the total 90 trials of simulation in CARLA.}
    \label{fig:fail_case}
\end{figure}
As shown in Fig.~\ref{fig:aggressiveOvertaking}, the EV overtakes the vehicle on the left aggressively and crashed it. One possible reason is that the prediction of other vehicles' trajectory is naive and the EV doesn't expect the surrounding vehicle to be so aggressive. 
In Fig.~\ref{fig:DrivingSolidLine}, the EV drives onto the solid yellow line, because of avoiding the vehicle on the right. One possible reason is that the PFs for lane markings are not in effect, because there is no explicit lane marking in the crossroad.

\section{Conclusions}
In this work, we propose an integrated behavior planning and motion control scheme for urban autonomous driving, in which we encode collision avoidance and compliance with common traffic rules as a unified potential field by soft constraints. High-level behavior planning is achieved with simultaneously low-level control. We implement the proposed framework on three challenging urban driving scenarios, where a high success rate and good tracking performance are attained. Moreover, we demonstrate the effectiveness of the designed PFs in generating safe and flexible maneuvers. 
Potential future work is to consider the trajectory prediction of the surrounding vehicles and develop guiding PFs in special areas such as intersections where explicit traffic regulations are not available.






\bibliographystyle{IEEEtran}
\bibliography{refs}

\begin{thebibliography}{10}
\providecommand{\url}[1]{#1}
\csname url@samestyle\endcsname
\providecommand{\newblock}{\relax}
\providecommand{\bibinfo}[2]{#2}
\providecommand{\BIBentrySTDinterwordspacing}{\spaceskip=0pt\relax}
\providecommand{\BIBentryALTinterwordstretchfactor}{4}
\providecommand{\BIBentryALTinterwordspacing}{\spaceskip=\fontdimen2\font plus
\BIBentryALTinterwordstretchfactor\fontdimen3\font minus
  \fontdimen4\font\relax}
\providecommand{\BIBforeignlanguage}[2]{{%
\expandafter\ifx\csname l@#1\endcsname\relax
\typeout{** WARNING: IEEEtran.bst: No hyphenation pattern has been}%
\typeout{** loaded for the language `#1'. Using the pattern for}%
\typeout{** the default language instead.}%
\else
\language=\csname l@#1\endcsname
\fi
#2}}
\providecommand{\BIBdecl}{\relax}
\BIBdecl

\bibitem{qi2022hierarchical}
Y.~Qi, B.~He, R.~Wang, L.~Wang, and Y.~Xu, ``Hierarchical motion planning for
  autonomous vehicles in unstructured dynamic environments,'' \emph{IEEE
  Robotics and Automation Letters}, vol.~8, no.~2, pp. 496--503, 2022.

\bibitem{prakash2021multi}
A.~Prakash, K.~Chitta, and A.~Geiger, ``Multi-modal fusion transformer for
  end-to-end autonomous driving,'' in \emph{Proceedings of the IEEE/CVF
  Conference on Computer Vision and Pattern Recognition}, 2021, pp. 7077--7087.

\bibitem{deng2019cooperative}
R.~Deng, B.~Di, and L.~Song, ``Cooperative collision avoidance for overtaking
  maneuvers in cellular {V2X}-based autonomous driving,'' \emph{IEEE
  Transactions on Vehicular Technology}, vol.~68, no.~5, pp. 4434--4446, 2019.

\bibitem{zhang2021enabling}
Z.~Zhang, L.~Zhang, J.~Deng, M.~Wang, Z.~Wang, and D.~Cao, ``An enabling
  trajectory planning scheme for lane change collision avoidance on highways,''
  \emph{IEEE Transactions on Intelligent Vehicles}, 2021.

\bibitem{chen2019model}
J.~Chen, B.~Yuan, and M.~Tomizuka, ``Model-free deep reinforcement learning for
  urban autonomous driving,'' in \emph{2019 IEEE Intelligent Transportation
  Systems Conference (ITSC)}.\hskip 1em plus 0.5em minus 0.4em\relax IEEE,
  2019, pp. 2765--2771.

\bibitem{ma2022alternating}
J.~Ma, Z.~Cheng, X.~Zhang, M.~Tomizuka, and T.~H. Lee, ``Alternating direction
  method of multipliers for constrained iterative {LQR} in autonomous
  driving,'' \emph{IEEE Transactions on Intelligent Transportation Systems},
  vol.~23, no.~12, pp. 23\,031--23\,042, 2022.

\bibitem{ma2022local}
J.~Ma, Z.~Cheng, X.~Zhang, Z.~Lin, F.~L. Lewis, and T.~H. Lee, ``Local learning
  enabled iterative linear quadratic regulator for constrained trajectory
  planning,'' \emph{IEEE Transactions on Neural Networks and Learning Systems},
  2022.

\bibitem{Huang2023decentralized}
Z.~Huang, S.~Shen, and J.~Ma, ``Decentralized {iLQR} for cooperative trajectory
  planning of connected autonomous vehicles via dual consensus {ADMM},''
  \emph{IEEE Transactions on Intelligent Transportation Systems}, pp. 1--13,
  2023.

\bibitem{guan2022integrated}
Y.~Guan, Y.~Ren, Q.~Sun, S.~E. Li, H.~Ma, J.~Duan, Y.~Dai, and B.~Cheng,
  ``Integrated decision and control: toward interpretable and computationally
  efficient driving intelligence,'' \emph{IEEE transactions on cybernetics},
  vol.~53, no.~2, pp. 859--873, 2022.

\bibitem{sahin2020autonomous}
Y.~E. Sahin, R.~Quirynen, and S.~Di~Cairano, ``Autonomous vehicle
  decision-making and monitoring based on signal temporal logic and
  mixed-integer programming,'' in \emph{2020 American Control Conference
  (ACC)}.\hskip 1em plus 0.5em minus 0.4em\relax IEEE, 2020, pp. 454--459.

\bibitem{eiras2021two}
F.~Eiras, M.~Hawasly, S.~V. Albrecht, and S.~Ramamoorthy, ``A two-stage
  optimization-based motion planner for safe urban driving,'' \emph{IEEE
  Transactions on Robotics}, vol.~38, no.~2, pp. 822--834, 2021.

\bibitem{rasekhipour2016potential}
Y.~Rasekhipour, A.~Khajepour, S.-K. Chen, and B.~Litkouhi, ``A potential
  field-based model predictive path-planning controller for autonomous road
  vehicles,'' \emph{IEEE Transactions on Intelligent Transportation Systems},
  vol.~18, no.~5, pp. 1255--1267, 2016.

\bibitem{wang2019crash}
H.~Wang, Y.~Huang, A.~Khajepour, Y.~Zhang, Y.~Rasekhipour, and D.~Cao, ``Crash
  mitigation in motion planning for autonomous vehicles,'' \emph{IEEE
  Transactions on Intelligent Transportation Systems}, vol.~20, no.~9, pp.
  3313--3323, 2019.

\bibitem{ma2019active}
Y.~Ma, P.~Zhang, and B.~Hu, ``Active lane-changing model of vehicle in b-type
  weaving region based on potential energy field theory,'' \emph{Physica A:
  Statistical Mechanics and its Applications}, vol. 535, p. 122291, 2019.

\bibitem{zhang2022integrated}
Q.~Zhang, L.~Qian, L.~Liu, L.~Ding, and F.~Yang, ``An integrated framework of
  autonomous vehicles based on distributed potential field in bev,'' in
  \emph{2022 IEEE International Conference on Robotics and Biomimetics
  (ROBIO)}.\hskip 1em plus 0.5em minus 0.4em\relax IEEE, 2022, pp. 1313--1320.

\bibitem{yan2022cooperative}
Y.~Yan, J.~Wang, Y.~Wang, C.~Hu, H.~Huang, and G.~Yin, ``A cooperative
  trajectory planning system based on the passengers' individual preferences of
  aggressiveness,'' \emph{IEEE Transactions on Vehicular Technology}, vol.~72,
  no.~1, pp. 395--406, 2022.

\bibitem{bae2019design}
S.~Bae, Y.~Kim, J.~Guanetti, F.~Borrelli, and S.~Moura, ``Design and
  implementation of ecological adaptive cruise control for autonomous driving
  with communication to traffic lights,'' in \emph{2019 American Control
  Conference (ACC)}.\hskip 1em plus 0.5em minus 0.4em\relax IEEE, 2019, pp.
  4628--4634.

\bibitem{ge2021numerically}
Q.~Ge, Q.~Sun, S.~E. Li, S.~Zheng, W.~Wu, and X.~Chen, ``Numerically stable
  dynamic bicycle model for discrete-time control,'' in \emph{2021 IEEE
  Intelligent Vehicles Symposium Workshops (IV Workshops)}.\hskip 1em plus
  0.5em minus 0.4em\relax IEEE, 2021, pp. 128--134.

\bibitem{Andersson2019}
J.~A.~E. Andersson, J.~Gillis, G.~Horn, J.~B. Rawlings, and M.~Diehl,
  ``{CasADi} -- {A} software framework for nonlinear optimization and optimal
  control,'' \emph{Mathematical Programming Computation}, vol.~11, no.~1, pp.
  1--36, 2019.

\bibitem{weerakoon2015artificial}
T.~Weerakoon, K.~Ishii, and A.~A.~F. Nassiraei, ``An artificial potential field
  based mobile robot navigation method to prevent from deadlock,''
  \emph{Journal of Artificial Intelligence and Soft Computing Research},
  vol.~5, no.~3, pp. 189--203, 2015.

\bibitem{dosovitskiy2017carla}
A.~Dosovitskiy, G.~Ros, F.~Codevilla, A.~Lopez, and V.~Koltun, ``{CARLA}: An
  open urban driving simulator,'' in \emph{Conference on Robot Learning}.\hskip
  1em plus 0.5em minus 0.4em\relax PMLR, 2017, pp. 1--16.

\end{thebibliography}

\end{document}